\documentclass[letterpaper]{article} 
\usepackage{aaai2026}  
\usepackage{times}  
\usepackage{helvet}  
\usepackage{courier}  
\usepackage[hyphens]{url}  
\usepackage{graphicx} 
\usepackage{amsmath}
\usepackage{amssymb}
\usepackage{xcolor}
\usepackage{booktabs}
\usepackage{multirow}
\usepackage{makecell}
\usepackage{graphicx}
\usepackage{subcaption}
\usepackage{dblfloatfix}
\usepackage{pifont}   
\usepackage{natbib}  
\usepackage{caption} 
\usepackage{algorithm}
\usepackage{algorithmic}

\usepackage{newfloat}
\usepackage{listings}
\DeclareCaptionStyle{ruled}{labelfont=normalfont,labelsep=colon,strut=off} 
\floatstyle{ruled}
\newfloat{listing}{tb}{lst}{}
\floatname{listing}{Listing}
\title{Towards Robust Semantic Correspondence: A Benchmark and Insights}
\author {
    Wenyue Chong\textsuperscript{\rm 1}
}
\affiliations {
    \textsuperscript{\rm 1}University of Chinese Academy of Sciences\\
    chongwenyue22@mails.ucas.ac.cn
}

\usepackage{bibentry}

\begin{document}

\maketitle

\begin{abstract}
Semantic correspondence aims to identify semantically meaningful relationships between different images and is a fundamental challenge in computer vision. It forms the foundation for numerous tasks such as 3D reconstruction, object tracking, and image editing. With the progress of large-scale vision models, semantic correspondence has achieved remarkable performance in controlled and high-quality conditions. However, the robustness of semantic correspondence in challenging scenarios is much less investigated. In this work, we establish a novel benchmark for evaluating semantic correspondence in adverse conditions. The benchmark dataset comprises 14 distinct challenging scenarios that reflect commonly encountered imaging issues, including geometric distortion, image blurring, digital artifacts, and environmental occlusion. Through extensive evaluations, we provide several key insights into the robustness of semantic correspondence approaches: (1) All existing methods suffer from noticeable performance drops under adverse conditions; (2) Using large-scale vision models can enhance overall robustness, but fine-tuning on these models leads to a decline in relative robustness; (3) The DINO model outperforms the Stable Diffusion in relative robustness, and their fusion achieves better absolute robustness; Moreover, We evaluate common robustness enhancement strategies for semantic correspondence and find that general data augmentations are ineffective, highlighting the need for task-specific designs. These results are consistent across both our dataset and real-world benchmarks.
\end{abstract}


\section{Introduction}

\begin{figure}[t]
    \centering
    \includegraphics[width=\linewidth]{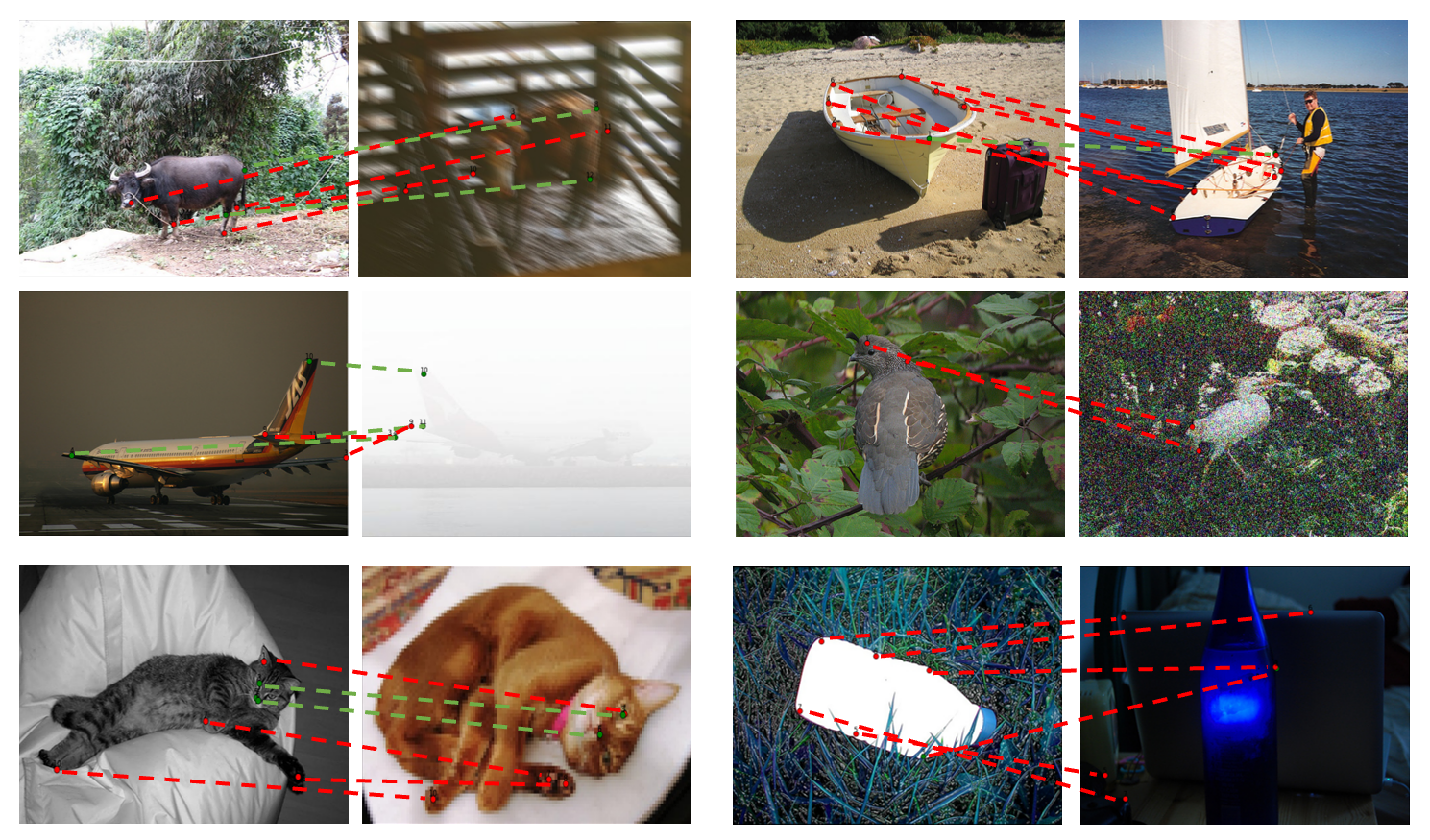}
    \caption{The state-of-the-art method fails at matching keypoints under adverse conditions. Red lines indicate incorrect correspondences, while green lines represent correctly established matches.}
    \label{fig:adverse_condition}
\end{figure}

Semantic correspondence is a core computer vision task that establishes pixel-level correspondences across instances within the same category. It serves as a crucial prerequisite for various applications such as 3D reconstruction~\cite{schonberger2016structure}, object tracking~\citep{gao2022aiatrack,wang2023tracking}, image and video editing~\citep{gupta2023asic,ofri2023neural,zhang2023tale,zhou2021transfill}. However, its real-world deployment remains challenging due to adverse imaging conditions. For example, extreme viewpoint variations induce severe geometric distortions that violate spatial consistency, and motion blur artifacts degrade local texture patterns essential for accurate matching. Figure~\ref{fig:adverse_condition} presents the performance of the leading semantic correspondence methods~\cite{zhang2024telling} under typical challenging conditions, showing that semantic correspondence remains unreliable and highlighting the importance of robust semantic correspondence.

With the advance of neural networks, deep learning-based semantic correspondence research has evolved through three key phases. Prior approaches employ supervised learning frameworks to learn accurate correspondences~\citep{cho2021cats,cho2022cats++,han2017scnet,jiang2021cotr,rocco2017convolutional,zeiler2014visualizing}, but their reliance on extensive ground-truth annotations limits their scalability. To address this issue, unsupervised methods are developed, which leverage the rich representations of Large-scale Vision Models~(LVMs), including DINO~\citep{caron2021emerging,oquab2023dinov2} and Stable Diffusion series~\cite{rombach2022high}, to establish direct pixel-to-pixel mappings. To improve task-specific adaptability, recent works propose supervised fine-tuning strategies on large-scale vision models through feature aggregation~\cite{luo2023diffusion}, conditional prompting~\cite{li2023sd4match}, and geometric alignment~\cite{zhang2024telling}, achieving state-of-the-art performance. Despite the above advancements, the robustness of semantic correspondence under adverse conditions is less explored. Such limitation arises from the absence of robustness benchmarks, as existing benchmarks, including SPair-71K~\cite{min2019spair} and PF-Pascal ~\cite{ham2016proposal}, emphasize intra-class appearance diversity rather than adverse conditions. Therefore, natural questions emerge: \textbf{(1) How do existing semantic correspondence methods perform under diverse adverse conditions? (2) If performance degradation occurs, what factors can enhance the robustness of semantic correspondence?}

In this paper, we propose a comprehensive benchmark for evaluating the robustness of semantic correspondence by introducing a novel benchmark dataset, Semantic Correspondence under Adverse Conditions~(SCAC), which comprises 14 adverse conditions across four key dimensions \textemdash Geometric variations, Blur\&Noise, Digital artifacts, and Environmental changes. To answer the questions above, we evaluate representative methods spanning three distinct paradigms on the SCAC dataset: supervised learning, unsupervised learning with LVMs, and supervised fine-tuning with LVMs. Through extensive experiments, we gain new insights into the behavior of semantic correspondence under adverse conditions: (1) All semantic correspondence methods show performance degradation under adverse conditions, especially for geometric variations. (2) Using LVMs enhances overall robustness, but fine-tuning with LVMs degrades relative robustness. (3) DINO exhibits greater relative robustness than Stable Diffusion, and their fusion obtains superior absolute robustness. We further explore whether widely-used robustness-enhancing strategies are applicable to semantic correspondence. Our findings highlight that existing robustness enhancement strategies for semantic correspondence tasks are insufficient, requiring tailored designs to improve performance across diverse adverse conditions.

In summary, our contributions can be summarized as follows:
\begin{itemize}
    \item To the best of our knowledge, this is the first attempt to introduce the benchmark for the robustness of semantic correspondence.
    \item We propose a new dataset tailored for assessing the robustness of semantic correspondence, encompassing a diverse range of prevalent adverse conditions to ensure a thorough evaluation.
    \item We offer novel insights into the robustness of semantic correspondence under adverse conditions, thereby directing future research toward robust semantic correspondence.
\end{itemize}

\section{Related Work}

\subsection{Semantic Correspondence Benchmarks}

Several benchmarks have been established to evaluate semantic correspondence methods. The PF-PASCAL~\cite{ham2016proposal} dataset, with 2,941 training pairs, 308 validation pairs, and 299 test pairs across 20 object categories, is constrained by its similar viewpoints and instance poses, limiting its ability to reflect the diverse perspectives encountered in real-world scenarios. Its supplement, PF-WILLOW~\cite{ham2017proposal}, provides an additional 900 test pairs but does not address this limitation. In contrast, SPair-71K~\cite{min2019spair} constructed from the Pascal VOC dataset~\cite{everingham2015pascal}, a more comprehensive benchmark, offers 53,340 training pairs, 5,384 validation pairs, and 12,234 test pairs across 18 object categories. It is designed to incorporate substantial scale variations and appearance diversity, tackling challenges such as viewpoint changes, scale differences, occlusions, and truncations. However, while SPair-71K addresses these four variations, it falls short of capturing the full spectrum of real-world complexities, such as digital artifacts and environmental changes. Furthermore, the lack of systematic robustness benchmarking means that the performance of existing methods in complex scenarios has not been adequately verified. This highlights the need for more comprehensive and robust evaluation frameworks to better assess the capabilities of semantic correspondence methods under diverse and challenging conditions.


\subsection{Semantic Correspondence Algorithms}

Semantic correspondence has evolved from hand-crafted descriptors~\citep{dalal2005histograms,lowe2004distinctive} to CNN-based alignment~\citep{han2017scnet,kim2017fcss,lee2019sfnet,rocco2017convolutional,seo2018attentive} and, to transformer architectures~\citep{cho2021cats,cho2022cats++,jiang2021cotr,kim2022transformatcher,zhao2021multi} by enabling global interaction capabilities. To mitigate the cost of dense annotations, the field now leverages large-scale pre-trained vision models. Self-supervised paradigms such as DINO~\citep{caron2021emerging,oquab2023dinov2} and Stable Diffusion~\citep{rombach2022high,tang2023emergent}learn correspondence without manual labels. These frozen backbones are further refined by lightweight fine-tuning: feature distillation ~\cite{fundel2025distilldift}, prompt-based adaptation ~\cite{li2023sd4match}, and geometry-aware tuning ~\cite{zhang2024telling}, achieving state-of-the-art accuracy on standard benchmarks.

\begin{figure*}[htbp]
  \centering
  \includegraphics[width=0.75\textwidth]{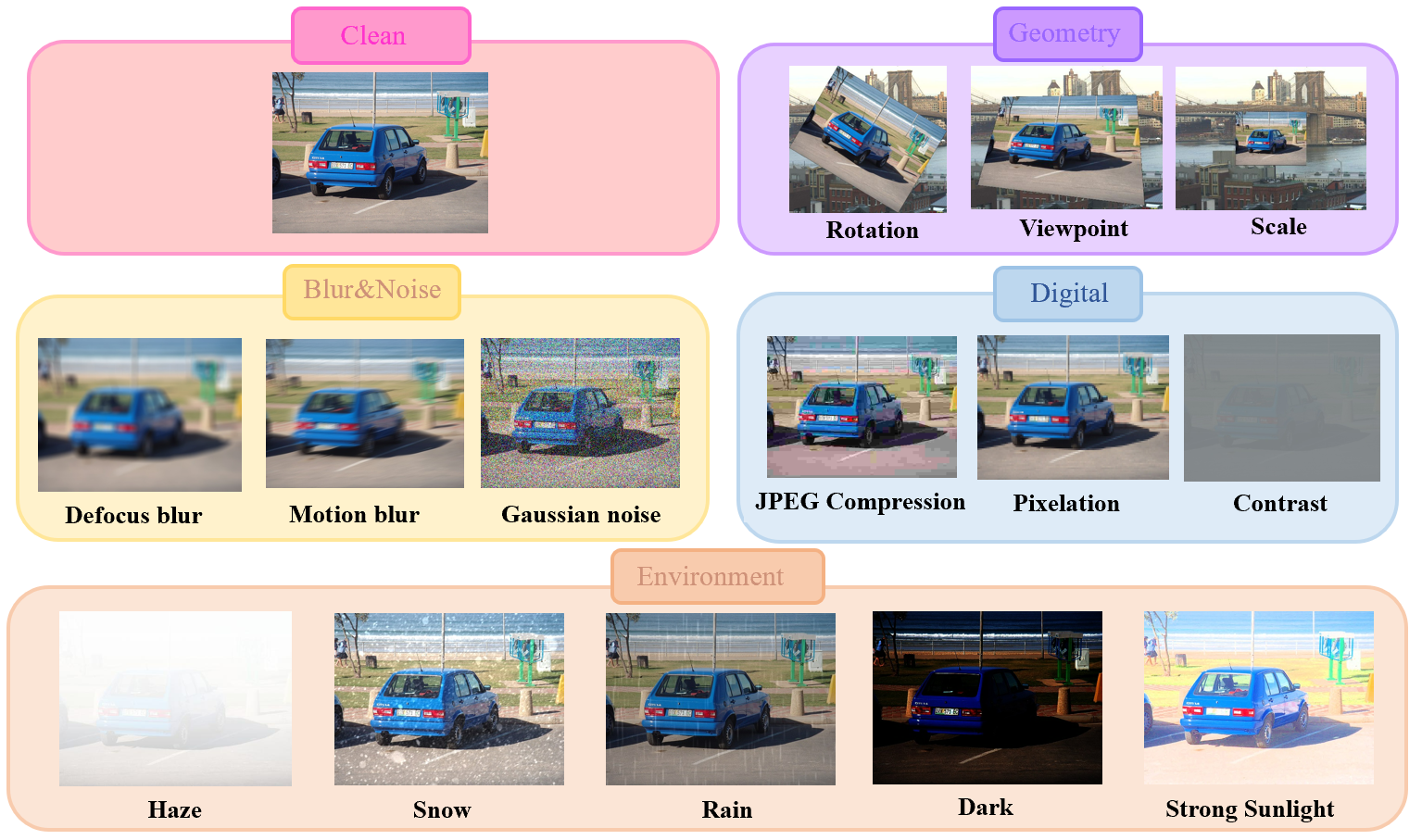}
  \caption{Visualization of the dataset construction. The figure presents the selected clean images alongside 14 challenging scenarios categorized into four distinct types.}
  \label{fig:数据集图片}
\end{figure*}
\section{Problem Setup}
\label{Problem Setup}

Our goal is to establish robust pixel-level correspondences across different visual instances within the same category under adverse conditions. Given a image pair $(x_i^s,x_i^t)$ sampled from the dataset $D=\{(x^s_{i},x^t_i)\}_{i=1}^{n}$ , $n$ is the sample size of $D$. Define $x_i^s$ as the source image and $x_i^t$ as the target image. Let $A=\{A_j\}^m_{j=1}$ be a family of functions, where each $A_j$ models image-level transformations caused by a specific adverse condition. For the target image $x^t_i$, the transformed image is $\hat{x}^t_i = A_j(x^t_i)$. The objective is to find a semantic correspondence mapping $f^* \colon \mathbf{p}^s \to \hat{\mathbf{p}}^t$ from the keypoint set $\mathbf{P}^s$ in $x_i^s$ to the corresponding set $\hat{\mathbf{P}}^t$ in $\hat{x}^t_i$:
\begin{equation}
    \min_f \mathbb{E}_{\substack{(\mathbf{x}^s, \mathbf{x}^t) \sim Q \\ A_j \sim A}} \left[ \mathrm{d}\left(\hat{\mathbf{p}}^t, f(\mathbf{p}^s)\right) \right]
\end{equation}
where $\mathrm{d}(\cdot,\cdot)$ denotes a distance metric (typically the Euclidean distance $||\cdot||_2$). Note that the functions $A$ are not accessible during the model training phase.



\section{Robustness Benchmark for Semantic Correspondence}
We provide a detailed overview of the robustness benchmark for evaluating semantic correspondence algorithms. First, we describe the construction of the benchmark dataset called SCAC. Next, we introduce the evaluation metrics used to measure robustness, including absolute robustness and relative robustness. Finally, we present the semantic correspondence methods selected for evaluation.

\subsection{Dataset Construction}
\label{Dataset Construction}
We propose a benchmark dataset called Semantic Correspondence in Adverse Conditions~(SCAC). The SCAC dataset includes 14 adverse conditions frequently encountered during image acquisition, categorized into four groups--Geometry, Blur\&Noise, Digital, and Environment--according to the nature of their visual effects on images. Note that the SCAC dataset, used exclusively for performance evaluation, does not participate in training.
\begin{figure*}[t]
\centering
\includegraphics[width=0.8\textwidth]{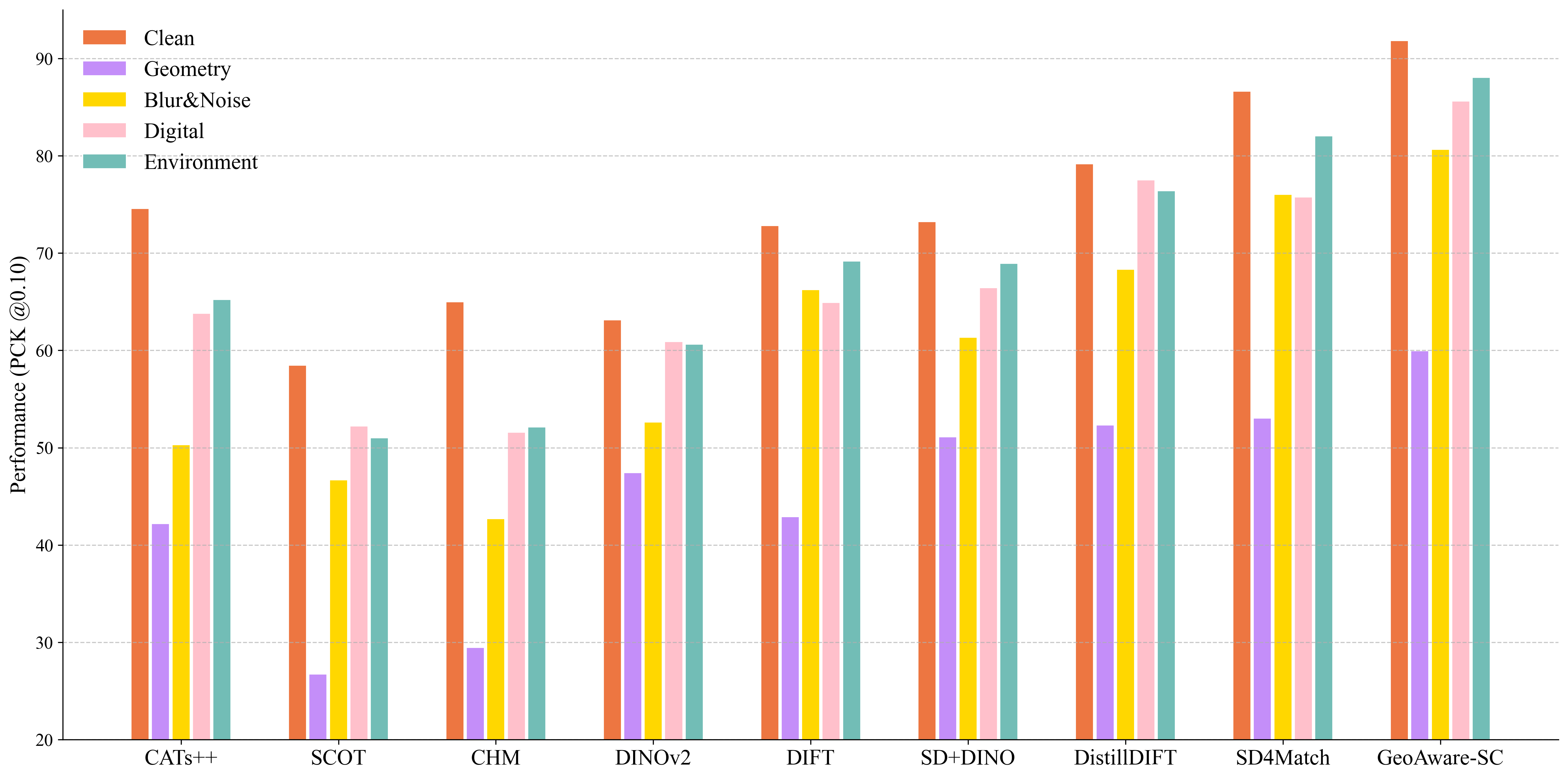}
\caption{Performance comparison on the SCAC dataset.
The bar chart shows the accuracy of different methods on our challenging benchmark. Higher bars reflect stronger robustness and reliability under adverse conditions.}
\label{fig:柱状图}
\end{figure*}
\subsubsection{The Acquisition of the Clean Subset}  
To establish a controlled baseline, we first construct a clean subset free of adverse conditions. The clean subset is used to (1) establish baseline performance for evaluating semantic correspondence, and (2) assess the performance gaps under various adverse condition types compared to clean data, serving as one of the robustness metrics. The clean subset is derived from the SPair-71K dataset, with samples affected by confounding factors--viewpoint shifts, scale variations, object truncation, and occlusion--filtered out. The refined subset ultimately consists of 1,464 pairs across 18 object categories, surpassing the other standard semantic correspondence datasets like PF-Willow~(900 testing pairs, 5 classes) and PF-Pascal~(299 testing pairs, 20 classes) in both scale and breadth of coverage. 

\subsubsection{The SCAC Dataset}
We simulate 14 common adverse conditions on the clean subset to systematically evaluate the robustness of semantic correspondence algorithms. In total, this results in $14 \times 1{,}464 = 20{,}496$ image pairs in the SCAC dataset. Based on their visual characteristics and underlying challenges, these adverse conditions are grouped into four categories, as illustrated in Figure~\ref{fig:数据集图片}:
(1) \textbf{Geometry.} Geometric transformations are essential to simulate spatial variations that often arise due to changes in object pose, camera viewpoint, or scale—factors that can severely affect semantic correspondence by altering the spatial configuration of object parts. To reflect these challenges, we consider three types of transformations: Rotation, which changes the global orientation of the object; Viewpoint, which simulates perspective distortion caused by viewing angle shifts; and Scale, which modifies the relative object size within the scene~\cite{liu2019gift}. These perturbations disrupt spatial consistency and test the model’s ability to preserve semantic alignment under structural variation. (2) \textbf{Blur\&Noise.} This category focuses on local degradations that weaken the fine-grained structures crucial for accurate correspondence. In practical scenarios, blur and noise often result from optical defocus, motion during capture, or sensor interference. We simulate three representative types: Defocus Blur, which smooths edges and textures; Motion Blur, which distorts local geometry through directional streaks; and Gaussian Noise, which adds random pixel-level fluctuations~\cite{hendrycks2019robustness}. These corruptions can significantly impair feature localization and matching precision. (3) \textbf{Digital.} Digital artifacts commonly emerge from compression and image resampling in real-world pipelines and can obscure or distort key visual signals. To model such effects, we include JPEG Compression, Pixelation, and Contrast variation. These corruptions simulate visual degradation such as compression noise, reduced spatial resolution, and global intensity shifts, respectively~\cite{hendrycks2019robustness}. Though not physically induced, they challenge the reliability of pixel-wise features and mid-level representations. (4) \textbf{Environment.} Dynamic environmental conditions—such as rain, fog, or extreme lighting—can significantly alter scene appearance and occlude semantic structures. We simulate five common adverse settings: Rain and Snow, which introduce partial occlusions and scattering effects; Fog, which reduces global visibility; Dark, which simulates underexposed scenes; and Strong Sunlight, which induces overexposure and contrast imbalance~\citep{hendrycks2019robustness,chen2025towards,chen2021all,li2018benchmarking}. These corruptions affect both global illumination and local clarity, posing challenges for extracting reliable features.

\subsection{Evaluation Metrics}
\label{Evaluation Metrics}
We use absolute robustness and relative robustness to
evaluate the robustness of semantic correspondence methods. Firstly, we employ Percentage of Correct Keypoints~(PCK), a generic metric for semantic correspondence. Formally, given a keypoint $\mathbf{p}_k^s$ in the source image $x^s_i$ and a keypoint $\mathbf{p}_k^t$ in the target image $x^t_i$, PCK is defined as:
\begin{equation}
\begin{split}
&\mathcal{PCK}(x_i^s, x_i^t) =\\
&\frac{1}{K}\sum_{k=1}^{K}
  \mathbb{I}\!\left(\,
    \|\mathbf{p}_k^t - f(\mathbf{p}_k^s)\|_2
    \leq \alpha \max(h_{\text{bbox}}, w_{\text{bbox}})
  \,\right)
\end{split}
\end{equation}\label{eq:PCK}
where $\alpha$ denotes an error tolerance threshold (typically set to 0.10), beyond which a keypoint correspondence is considered to be invalid. $h_{bbox}, w_{bbox}$ denote the height and width of the bounding box enclosing the object in the target image $x_i^t$. The function $\mathbb{I}(\cdot)$ is a binary indicator function with $\mathbb{I}(\text{True}) = 1$ and $\mathbb{I}(\text{False}) = 0$. $K$ represents the number of keypoints. Higher PCK values indicate better semantic correspondence performance. In the following sections, PCK@0.1 denotes the value when $\alpha$ is set to 0.1. 

Given sample pairs $\{(x^s_{i},\hat{x}^t_i)\}_{i=1}^{N}$, where $\hat{x}^t_i$ represents the image generated under the adverse condition $A_j$, the absolute robustness under $A_j$ can be calculated as:
\begin{equation}
    \text{Absolute Robustness}:= \frac{1}{N} \sum_{i=1}^{N}\mathcal{PCK}(x_i^s, \hat{x}^t_i)
\end{equation}\label{eq:absolute robustness}
where $N$ denotes the number of sample pairs. Measuring absolute robustness alone is insufficient to fully capture the performance of semantic correspondence under adverse conditions. For example, Method\#1~(M1) achieves 73\% PCK on clean scenarios and 50\% under adverse conditions, while Method\#2~(M2) scores 67\% on clean scenarios and 47\% under the same conditions. Although M1 has better absolute robustness~(50\% vs. 47\%), M2 shows a smaller performance drop between clean and adverse conditions (20\% vs. 23\%). To this end, we define the relative robustness of the function $f$ as:
\begin{equation}
    \begin{split}
    &\text{Relative Robustness}:= \\
    &1-(\frac{1}{N} \sum_{i=1}^{N}\mathcal{PCK}(x_i^s, x^t_i)-\frac{1}{N} \sum_{i=1}^{N}\mathcal{PCK}(x_i^s, \hat{x}^t_i)) 
    \end{split}
\end{equation}\label{eq:relative robustness}
Therefore, we adopt absolute robustness and relative robustness jointly evaluate robustness under
adverse conditions.
\begin{figure*}[htbp]
  \centering
  \includegraphics[width=0.9\linewidth]{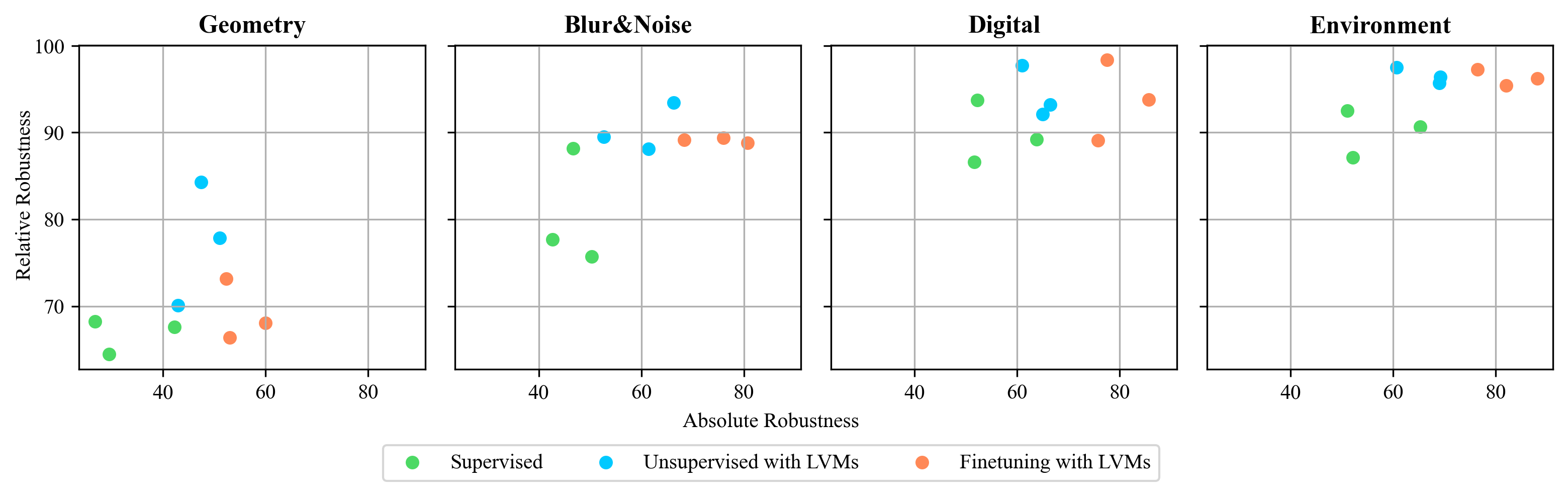}
  \caption{Absolute and relative robustness analysis of three categories of methods across four different scenarios. In the scatter plot, higher x-values indicate greater absolute robustness, while higher y-values represent stronger relative robustness across varying conditions.}
  \label{fig:scatter_plot}
\end{figure*}
\subsection{Evaluation Methods}
\label{Comparison Methods}
We select eleven representative methods from three distinct learning paradigms and conduct comprehensive evaluations on the SCAC dataset. In supervised learning, we evaluate three methods specializing in distinct areas: CATs++~\cite{cho2022cats++} in cost aggregation optimization, SCOT~\cite{liu2020semantic} in optimal transport modeling, and CHM~\cite{min2021chm} in network architecture modification. Notably, CATs++ achieves SOTA performance within this paradigm. For unsupervised learning with LVMs, we choose DINOv1~\cite{caron2021emerging} and DINOv2~\cite{oquab2023dinov2}, as well as Stable Diffusion v1.5~\cite{rombach2022high} and Stable Diffusion v2.1~\cite{tang2023emergent}, all of which have demonstrated effectiveness in semantic correspondence tasks. Moreover, we include the SD+DINO approach~\cite{zhang2023tale}, which merges the structural awareness of generative models with the semantic consistency of discriminative models. For finetuning with LVMs, we evaluate three methods: DistillDIFT~\cite{fundel2025distilldift}, SD4Match~\cite{li2023sd4match}, and GeoAware-SC~\cite{zhang2024telling}. These approaches utilize supervised fine-tuning on LVMs via knowledge distillation, conditional prompting, and geometry-aware constraints, respectively, achieving superior performance in semantic  correspondence tasks.

\section{Experiments And Insights}
\subsection{How do various types of adverse conditions affect overall correspondence accuracy?} 
Figure~\ref{fig:柱状图} shows the semantic correspondence performance of nine methods on the clean subset and four adverse conditions: Geometry, Blur\&Noise, Digital, and Environment. The PCK values represent average accuracy within each category (e.g., Geometry averages Rotation, Viewpoint, and Scale). Detailed results are in the supplementary material. As seen in Figure~\ref{fig:柱状图}, methods are most sensitive to geometric changes, with SCOT and CHM dropping to around 30\% PCK. GeoAware-SC also declines sharply from over 90\% on clean data to below 60\% under geometric distortions. Blur\&Noise causes the second largest accuracy drop for most methods, where CATs++, SCOT, and CHM achieve only 40–55\% PCK. In contrast, Digital and Environment corruptions have the least impact, differing from other vision tasks where they usually cause more severe degradation~\citep{dalva2023benchmarking,dong2023benchmarking,hendrycks2019benchmarking,li2023common}. This reveals unique vulnerability patterns in semantic correspondence. Overall, these findings demonstrate that \textbf{all evaluated methods suffer varying degrees of performance loss under the four major adverse conditions}, underscoring the urgent need to improve robustness in semantic correspondence frameworks.
\begin{table*}[t]
\centering

\resizebox{\textwidth}{!}{
\begin{tabular}{@{}
    l
    c
    *{10}{c}
    @{}}
\toprule
\multirow{2}{*}{\textbf{Method} \rule{0pt}{12pt}} & 
\multirow{2}{*}{\textbf{Clean}} & 
\multicolumn{5}{c}{\textbf{Absolute Robustness}} & 
\multicolumn{5}{c}{\textbf{Relative Robustness}} \\
\cmidrule(lr){3-7} \cmidrule(l){8-12}
& & Geometry & Blur\&Noise & Digital & Environment & Avg & Geometry & Blur\&Noise & Digital & Environment & Avg \\
\midrule
\textbf{DINOv1}   & 46.10 & 24.38 & 37.19 & 40.45 & 39.13 & 35.29 & \underline{78.28} & \underline{91.09} & \underline{94.35} & 93.03 & \underline{89.19} \rule{0pt}{11pt} \\
\textbf{DINOv2}   & 63.17 & \underline{47.45} & 52.66 & 60.91 & 60.65 & 55.42 & \textbf{84.28} & 89.49 & \textbf{97.74} & \textbf{97.48} & \textbf{92.25} \\
\cmidrule(lr){1-12}
\textbf{SDv1.5}   & 68.71 & 41.04 & 51.55 & 49.95 & 61.31 & 50.96 & 72.33 & 82.84 & 81.24 & 92.60 & 82.25 \\
\textbf{SDv2.1}   & \underline{72.84} & 42.94 & \textbf{66.27} & \underline{64.95} & \textbf{69.21} & \underline{60.84} & 70.10 & \textbf{93.43} & 92.11 & \underline{96.37} & 88.00 \\
\cmidrule(lr){1-12}
\textbf{SD+DINO}  & \textbf{73.26} & \textbf{51.13} & \underline{61.37} & \textbf{66.46} & \underline{68.97} & \textbf{61.98} & 77.87 & 88.11 & 93.20 & 95.71 & 88.72 \\
\bottomrule
\end{tabular}
}
\caption{Performance comparison of the DINO series, Stable Diffusion series, and their combination (SD+DINO) across four challenging scenarios in the SCAC dataset: \textit{Geometry}, \textit{Blur \& Noise}, \textit{Digital}, and \textit{Environment}. Both absolute and relative robustness are reported. For each scenario, the best-performing score is highlighted in \textbf{bold}, and the second-best is \underline{underline}.}
\label{tab:comparison}
\end{table*}




\begin{table}[t]
\centering
\resizebox{1.0\columnwidth}{!}{
\Large
\begin{tabular}{@{}l ccccc@{}}
\toprule
\textbf{Method} & \multicolumn{5}{c}{\textbf{CKA}} \\
\cmidrule(lr){2-6}
& Geometry & Blur\&Noise & Digital & Environment & Avg \\
\midrule
\textbf{DINOv1}   & \underline{0.9655} & \underline{0.9788} & \underline{0.9845} & 0.9699 & \underline{0.9747} \\
\textbf{DINOv2}   & \textbf{0.9800} & 0.9767 & \textbf{0.9914} & \textbf{0.9939} & \textbf{0.9855} \\
\cmidrule(lr){1-6}
\textbf{SDv1.5}   & 0.8873 & 0.8909 & 0.8835 & 0.8939 & 0.8889 \\
\textbf{SDv2.1}   & 0.8145 & \textbf{0.9817} & 0.9626 & \underline{0.9832} & 0.9355 \\
\cmidrule(lr){1-6}
\textbf{SD+DINO}  & 0.9467 & 0.9639 & 0.9766 & 0.9807 & 0.9669 \\
\bottomrule
\end{tabular}
}
\caption{CKA similarity scores between clean and perturbed image features under four adverse conditions. Higher is better. Best and second-best scores are in \textbf{bold} and \underline{underline}, respectively.}
\label{tab:comparison_CKA}
\end{table}

\subsection{How Does the Learning Paradigm Influence the Robustness?} 


The scatter plots in Figure~\ref{fig:scatter_plot} compare the robustness of semantic correspondence methods across three paradigms—Supervised learning (green markers), Unsupervised learning with LVMs (blue markers), and Finetuning with LVMs (orange markers)—under four types of adverse conditions: Geometry, Digital, Blur\&Noise, and Environment. The x-axis indicates Absolute Robustness, while the y-axis shows Relative Robustness. Distinct marker clusters reveal the robustness profiles of each paradigm, with detailed plots available in the supplementary material.

Supervised learning shows the weakest robustness overall, with green markers clustered in the lower-left, reflecting poor generalization due to limited labeled data. In Geometry, it achieves only 25\%-45\% absolute and 60\%-70\% relative robustness; similarly low values are observed under Blur\&Noise. These findings highlight the limitations of supervised learning in maintaining semantic correspondence under adverse conditions.

Unsupervised learning with LVMs offers clear improvements, shifting blue markers upward and rightward. This stems from LVMs' self-supervised pretraining on vast unlabeled data, encoding structural priors and global context that enable better generalization to distortions. Finetuning with LVMs further boosts absolute robustness (e.g., up to 82\% in Blur\&Noise), with orange markers extending far right. However, its relative robustness often declines compared to unsupervised models, causing its markers to sit lower on the y-axis than blue markers despite a rightward shift. These results reveal two key insights: \textbf{(1) the integration of LVMs substantially boosts both absolute and relative robustness across all adverse conditions; (2) while finetuning enhances absolute robustness, it fails to improve relative robustness and even exhibits performance degradation in challenging conditions like Geometry and Blur\&Noise.}

\subsection{Which LVM is More Robust: DINO or Stable Diffusion?}

The above results confirm the overall robustness advantage of LVM-based methods. As LVMs become widely adopted in semantic correspondence, it is essential to identify which specific models are most robust. We evaluate several popular LVMs—DINOv1, DINOv2, SDv1.5, SDv2.1, and their combination (SD+DINO)—all of which have shown strong performance in prior work. As shown in Table~\ref{tab:comparison}, SDv2.1 achieves the highest absolute robustness (60.84\% on average), except in Geometry, where DINOv2 leads by over 4\%. In contrast, DINO models consistently deliver higher relative robustness (92.25\% for DINOv2 vs. 88.00\% for SDv2.1). Their fusion (SD+DINO) further improves absolute robustness to 61.98\%, but slightly lags DINOv2 in relative terms. \textbf{These results suggest that
Stable Diffusion offers greater absolute robustness; DINO models are preferable for maintaining relative robustness.} 

\begin{figure}[htbp]
  \centering
  \includegraphics[width=\linewidth]{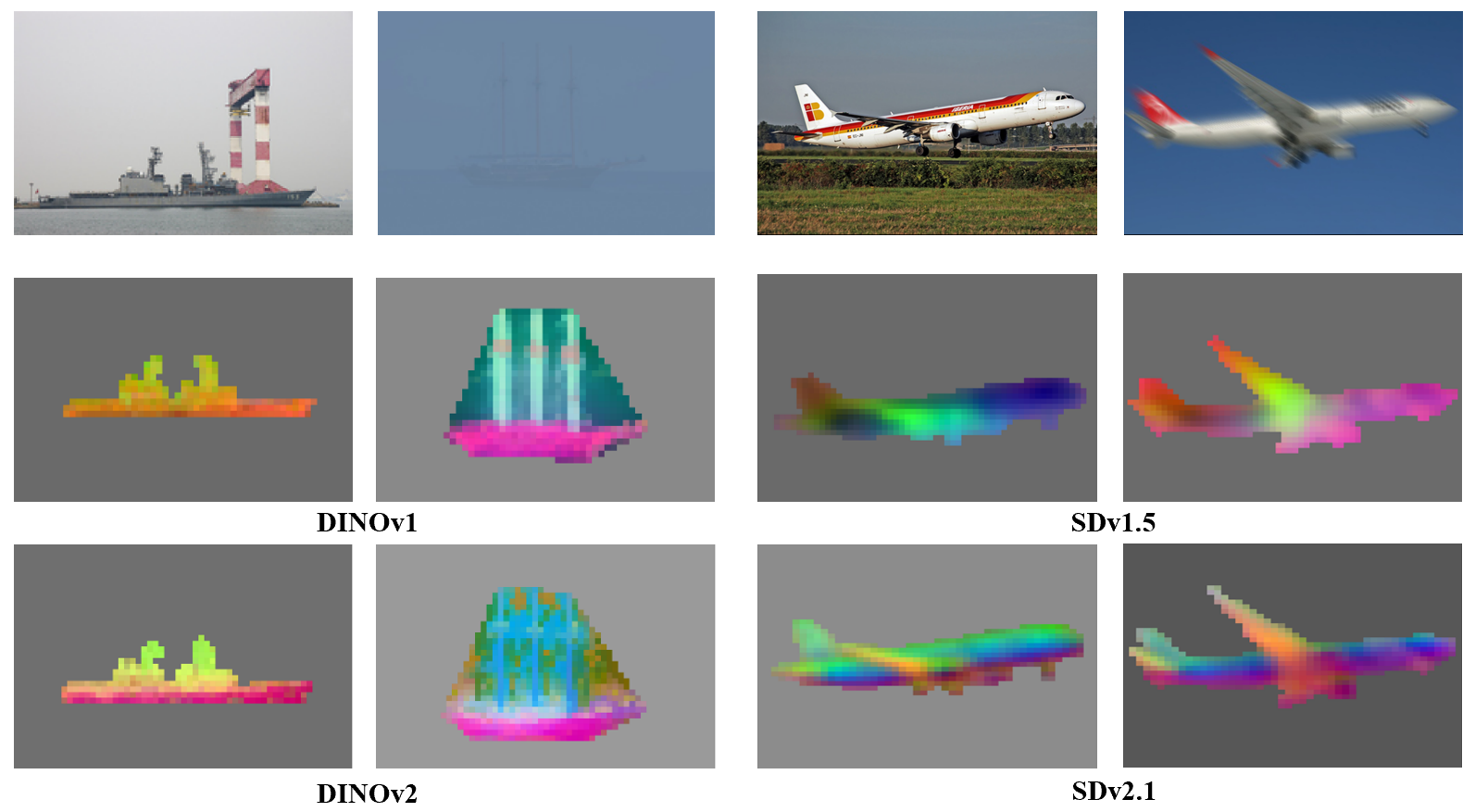}
  \caption{Feature maps extracted by DINOv1/v2 and SDv1.5/v2.1 on the image pairs.}
  \label{fig:pca_feats}
\end{figure}

To better understand why DINO models achieve higher relative robustness under adverse conditions, we conduct a Centered Kernel Alignment (CKA)~\cite{kornblith2019similarity} analysis between clean and perturbed versions of the same image. CKA quantifies the similarity of feature representations from two inputs by comparing their pairwise feature correlations. A higher CKA value implies greater preservation of semantic structure under perturbation, which directly aligns with the concept of relative robustness—i.e., a model's ability to maintain internal consistency despite external corruption. As shown in Table~\ref{tab:comparison_CKA}, DINOv2 consistently yields higher CKA scores across various adverse conditions, mirroring its superior relative robustness metrics and suggesting stronger invariance to noise and distortion at the representation level. Within each series, newer versions show clear improvements: DINOv2 outperforms DINOv1 in both absolute (35.29\% $\rightarrow$ 55.42\%) and relative robustness (89.19\% $\rightarrow$ 92.25\%) due to stronger pretraining and architectural upgrades; SDv2.1 also surpasses SDv1.5 (50.96\% $\rightarrow$ 60.84\% absolute, 82.25\% $\rightarrow$ 88.00\% relative) through better data and model refinement. These trends are visually supported in Figures~\ref{fig:pca_feats}, where DINOv2 and SDv2.1 produce more semantically meaningful feature maps than their earlier counterparts.

\begin{table*}[htbp]
\centering

\resizebox{\textwidth}{!}{
\huge
\begin{tabular}{@{} 
    l 
    c  
    *{5}{c} 
    *{5}{c} 
    @{}} 
\toprule
\multirow{2}{*}{\textbf{Method}} & 
\multirow{2}{*}{\textbf{Clean}} & 
\multicolumn{5}{c}{\textbf{Absolute Robustness}} & 
\multicolumn{5}{c}{\textbf{Relative Robustness}} \\
\cmidrule(lr){3-7} \cmidrule(l){8-12}
& & Geometry & Blur\&Noise & Digital & Environment & Avg & Geometry & Blur\&Noise & Digital & Environment & Avg \\
\midrule
\textbf{Baseline} & 
77.16 & 
43.23 & 
60.97 & 
60.05 & 
71.39 & 
58.91 & 
66.07 & 
83.81 & 
82.89 & 
94.23 & 
81.75 \\

\cmidrule(lr){1-12} 

\textbf{RandAug} & 
74.53{(-)} & 
38.08{(-)} & 
52.59{(-)} & 
53.93{(-)} & 
68.02{(-)} & 
53.15{(-)} & 
63.55{(-)} & 
78.06{(-)} & 
79.40{(-)} & 
93.49{(-)} & 
78.62{(-)} \\

\textbf{AutoAug} & 
76.45{(-)} & 
40.00{(-)} & 
57.74{(-)} & 
57.30{(-)} & 
69.89{(-)} & 
56.23{(-)} & 
63.55{(-)} & 
81.29{(-)} & 
80.85{(-)} & 
93.44{(-)} & 
79.78{(-)} \\

\textbf{AugMix} & 
77.45{(+)} & 
42.25{(-)} & 
60.84{(-)} & 
59.31{(-)} & 
71.03{(-)} & 
58.36{(-)} & 
64.80{(-)} & 
83.39{(-)} & 
81.86{(-)} & 
93.58{(-)} & 
80.91{(-)} \\

\cmidrule(lr){1-12} 

\textbf{AugMix*} & 
77.00{(-)} & 
43.11{(-)} & 
61.57{(+)} & 
60.54{(+)} & 
71.41{(+)} & 
59.16{(+)} & 
66.11{(+)} & 
\underline{84.57}{(+)} & 
\underline{83.54}{(+)} & 
\underline{94.41}{(+)} & 
\underline{82.16}{(+)} \\

\textbf{AugMix*+CL} & 
77.77{(+)} & 
43.10{(-)} & 
\textbf{62.78}{(+)} & 
\textbf{61.84}{(+)} & 
72.44{(+)} & 
60.04{(+)} & 
65.33{(-)} & 
\textbf{85.01}{(+)} & 
\textbf{84.07}{(+)} & 
\textbf{94.67}{(+)} & 
\textbf{82.27}{(+)} \\

\textbf{GeoAug} & 
78.04{(+)} & 
46.47{(+)} & 
59.48{(-)} & 
59.08{(-)} & 
71.51{(+)} & 
59.14{(+)} & 
\textbf{68.43}{(+)} & 
81.44{(-)} & 
81.04{(-)} & 
93.47{(-)} & 
81.10{(-)} \\

\cmidrule(lr){1-12} 

\textbf{AugMix*+GeoAug} & 
\underline{78.81}{(+)} & 
\textbf{47.14}{(+)} & 
61.31{(+)} & 
\underline{61.21}{(+)} & 
\underline{73.00}{(+)} & 
\underline{60.67}{(+)} & 
\underline{68.33}{(+)} & 
82.50{(-)} & 
82.40{(-)} & 
94.19{(-)} & 
81.86{(+)} \\

\textbf{DC via AugMix*+GeoAug} & 
\textbf{79.40}{(+)} & 
\underline{46.94}{(+)} & 
\underline{62.05}{(+)} & 
61.09{(+)} & 
\textbf{73.21}{(+)} & 
\textbf{60.82}{(+)} & 
67.54{(+)} & 
82.65{(-)} & 
81.69{(-)} & 
93.81{(-)} & 
81.42{(-)} \\

\bottomrule
\end{tabular}
}
\caption{
Performance comparison of various data augmentation strategies under clean and perturbed conditions.
(+) highlights improvements over the baseline, while (-) indicates a performance degradation.
Bold numbers highlight the best-performing method in each setting, and underlined values indicate the second-best.
}
\label{tab:comparison_re}
\end{table*}


\subsection{Do Robustness-Enhancing Methods Effectively Improve the Robustness of Semantic Correspondence Models?}
To improve robustness, we evaluate existing techniques for semantic correspondence, grouped into three types, these methods are used to augment the training data, followed by model training: (1) Composite data augmentations: We use AugMix~\cite{hendrycks2020augmix}, RandAugment~\cite{cubuk2020randaugment}, and AutoAugment~\cite{cubuk2019autoaugment}, which apply random image transformations and have proven effective in robust vision tasks~\citep{schneider2020improving,shu2022test,du2023dream,he2022masked,hassani2023neighborhood}. (2) Geometric and Non-Geometric augmentation strategies: To avoid corrupting ground truth labels, we design GeoAug inspired by~\cite{liu2019gift}, a geometric augmentation method that adjusts both images and their corresponding annotations. We also create AugMix* by removing geometric transforms from AugMix as a non-geometric variant(AugMix*+CL). Moreover, we incorporate the consistency regularization~\citep{sohn2020fixmatch,berthelot2019mixmatch,fan2023revisiting}, a widely adopted technique in robustness tasks. (3) The fusion strategy of Geometric and Non-Geometric augmentation: We test two fusion methods — a cascaded pipeline applying AugMix* then (GeoAugAugMix*+GeoAug), and offline dataset expansion combining separately augmented sets with the original data(DC via AugMix*+GeoAug). As stable diffusion models are widely used in semantic correspondence, we select SDv1.5 as our baseline.

Table~\ref{tab:comparison_re} shows the effectiveness of robustness-enhancing methods on our dataset. Composite data augmentations often hurt absolute robustness, especially in Blur\&Noise (e.g., RandAug 52.59\%, AutoAug 57.74\% vs. baseline 60.97\%). Single geometric or non-geometric augmentations improve robustness on average but vary by condition: GeoAug helps Geometry (46.47\% vs. 43.23\%) but worsens Blur\&Noise and Digital, while AugMix* improves Blur\&Noise and Digital but lowers Geometry (43.11\%). Adding consistency regularization (AugMix*+CL) further boosts robustness (60.04\% absolute, 82.27\% relative). The fusion method (DC via AugMix*+GeoAug) achieves the highest absolute robustness (60.82\%) but reduces relative robustness.
\subsection{Are Robustness-Enhancing Strategies Effective Under Real-World Conditions?}
A central question is whether robustness-enhancing methods that improve performance on synthetic distortions also remain effective under natural real-world corruptions. To assess this, we evaluated the same models on both the SCAC Dataset (with controlled synthetic corruptions) and a subset of SPair-71k containing naturally corrupted samples. As shown in Table~\ref{tab:consistency}, our proposed method—DC via AugMix+GeoAug*—achieves the best performance on both datasets, followed closely by AugMix+GeoAug*, demonstrating consistent improvements across distortion types. This consistency highlights the strong generalization ability of our approach and supports the SCAC Dataset as a reliable robustness benchmark. It is worth noting that, due to the lack of clean counterparts for naturally corrupted samples, only absolute robustness can be evaluated in real-world settings.


\begin{table}[h]
  \centering
  \small
  \resizebox{0.40\textwidth}{!}{\begin{tabular}{@{}lc@{}}
    \toprule
    \textbf{Method} & 
    \makecell{\textbf{Absolute Robustness} \\ \textbf{in Real-World Scenarios}} \\
    \midrule
    Baseline               & 62.75 \\
    RandAug                & 62.01 \\
    AutoAug                & 57.53 \\
    AugMix                 & 61.02 \\
    AugMix*                & 62.13 \\
    AugMix*+CL             & 63.08 \\
    GeoAug                 & 62.46 \\
    AugMix*+GeoAug         & \underline{64.34} \\
    DC via AugMix*+GeoAug  & \textbf{64.54} \\
    \bottomrule
  \end{tabular}}
  \caption{Performance of different data augmentation methods on real-world distorted datasets. Best and second-best scores are in bold and underline,
respectively.}
  \label{tab:consistency}
\end{table}


\section{Conclusion}
In this paper, we establish a pioneering benchmark to evaluate the robustness of semantic correspondence under adverse conditions, revealing critical insights into the performance of existing methods. We demonstrate that under adverse conditions, all existing methods exhibit substantial performance declines. While large-scale vision models enhance overall robustness, fine-tuning often compromises relative robustness, with DINO outperforming Stable Diffusion in relative robustness and their fusion achieving superior absolute robustness. Our findings underscore the limitations of robustness enhancement strategies for semantic correspondence. This benchmark and analysis pave the way for future advancements in robust semantic correspondence with significant implications for real-world applications.

\bibliography{aaai2026}
\end{document}